\documentclass[letterpaper, 10 pt, conference]{ieeeconf}  

\usepackage{cite}
\usepackage{amsmath,amssymb,amsfonts}
\usepackage{algorithmic}
\usepackage{graphicx}
\usepackage{textcomp}
\usepackage{xcolor}
\usepackage{amsmath} 
\usepackage{amssymb}  
\usepackage{comment}
\usepackage{siunitx}
\usepackage{subcaption}
\usepackage{multirow}
\usepackage{hyperref}
\usepackage{booktabs}
\usepackage{makecell}
\usepackage{soul}
\usepackage{threeparttable}

\IEEEoverridecommandlockouts                              

\title{\LARGE \bf
VOOM: Robust Visual Object Odometry and Mapping using Hierarchical Landmarks
}

\author{Yutong Wang$^{1}$, Chaoyang Jiang$^{1}$, Xieyuanli Chen$^{2}$
\thanks{*This work was supported by the National Key Research and Development Project of China (No. 2020YFC1512503), Fund for key Laboratory of Space Flight Dynamics Technology (Num 2022-JYAPAF-F1028), and Young Elite Scientists Sponsorship Program by CAST (No. 2023QNRC001). (Corresponding authors: Chaoyang Jiang and Xieyuanli Chen.)}
\thanks{$^{1}$Y. Wang and C. Jiang are with the School of Mechanical Engineering, Beijing Institute of Technology, China, and the Yangtze Delta Region Academy of Beijing Institute of Technology, Jiaxing, China.
        {\tt\small yutongwang@bit.edu.cn, cjiang@bit.edu.cn}}%
\thanks{$^{2}$X. Chen is with the College of Intelligence Science and Technology, National University of Defense Technology, China
        {\tt\small xieyuanli.chen@nudt.edu.cn}}%
}

\begin{document}

\maketitle
\thispagestyle{empty}
\pagestyle{empty}

\begin{abstract}

In recent years, object-oriented simultaneous localization and mapping (SLAM) has attracted increasing attention due to its ability to provide high-level semantic information while maintaining computational efficiency. Some researchers have attempted to enhance localization accuracy by integrating the modeled object residuals into bundle adjustment. However, few have demonstrated better results than feature-based visual SLAM systems, as the generic coarse object models, such as cuboids or ellipsoids, are less accurate than feature points. In this paper, we propose a Visual Object Odometry and Mapping framework VOOM using high-level objects and low-level points as the hierarchical landmarks in a coarse-to-fine manner instead of directly using object residuals in bundle adjustment. Firstly, we introduce an improved observation model and a novel data association method for dual quadrics, employed to represent physical objects. It facilitates the creation of a 3D map that closely reflects reality. Next, we use object information to enhance the data association of feature points and consequently update the map. In the visual object odometry backend, the updated map is employed to further optimize the camera pose and the objects. Meanwhile, local bundle adjustment is performed utilizing the objects and points-based covisibility graphs in our visual object mapping process. Experiments show that VOOM outperforms both object-oriented SLAM and feature points SLAM systems such as ORB-SLAM2 in terms of localization. The implementation of our method is available at \url{https://github.com/yutongwangBIT/VOOM.git}.

\end{abstract}

\section{Introduction}

Accurate localization and mapping are essential in various applications, including autonomous navigation, robot manipulation, and augmented reality. Cameras are cost-effective and easy to integrate while offering rich geometry and semantic information, making them practical for real-world scenarios. Consequently, visual odometry or simultaneous localization and mapping (SLAM) has attracted increasing attention from researchers.

\begin{figure}[th]
	\centering
	\begin{subfigure}[t]{0.48\textwidth}
		\includegraphics[width=0.92\linewidth, height=0.38\linewidth]{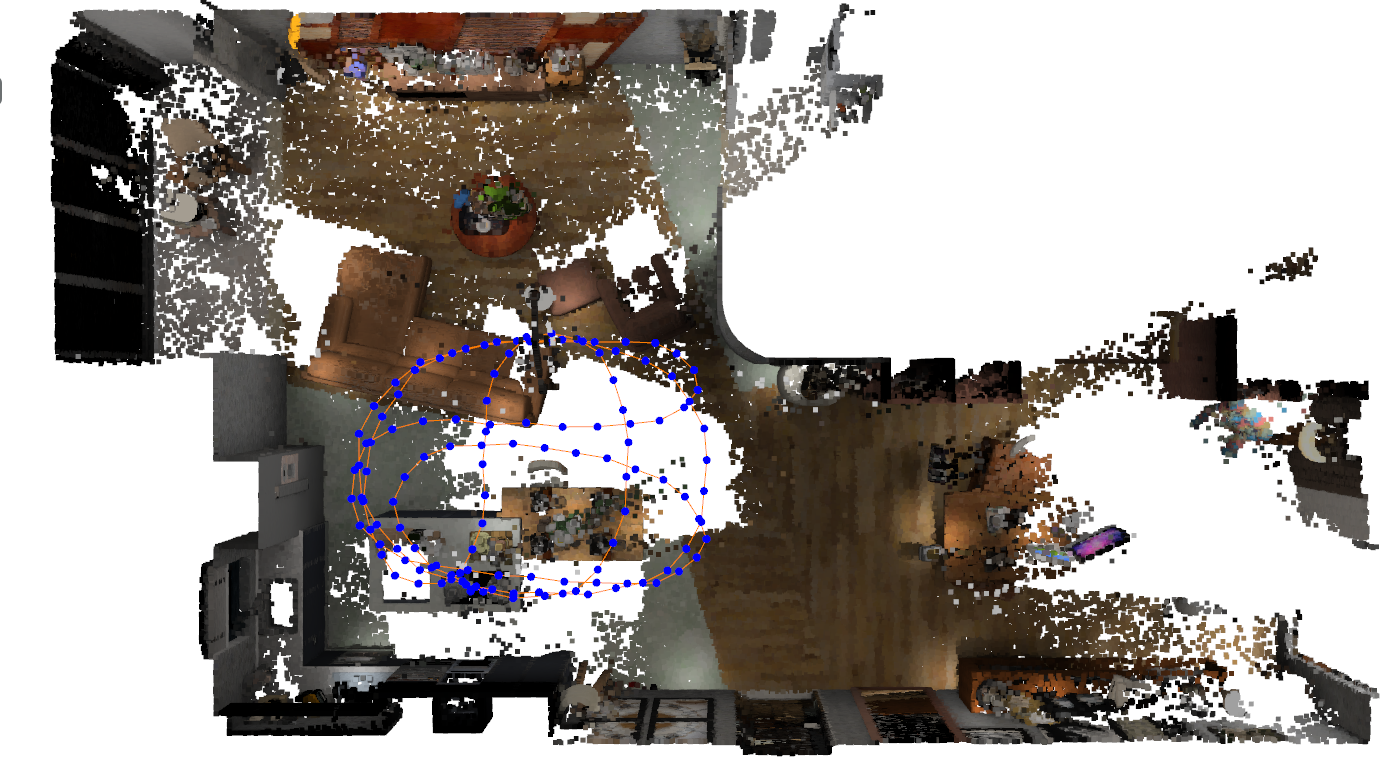}
		\label{fig:intro_a}
	\end{subfigure} \\ \vspace{2pt}
	\begin{subfigure}[t]{0.48\textwidth}
		\includegraphics[width=0.99\linewidth, height=0.38\linewidth]{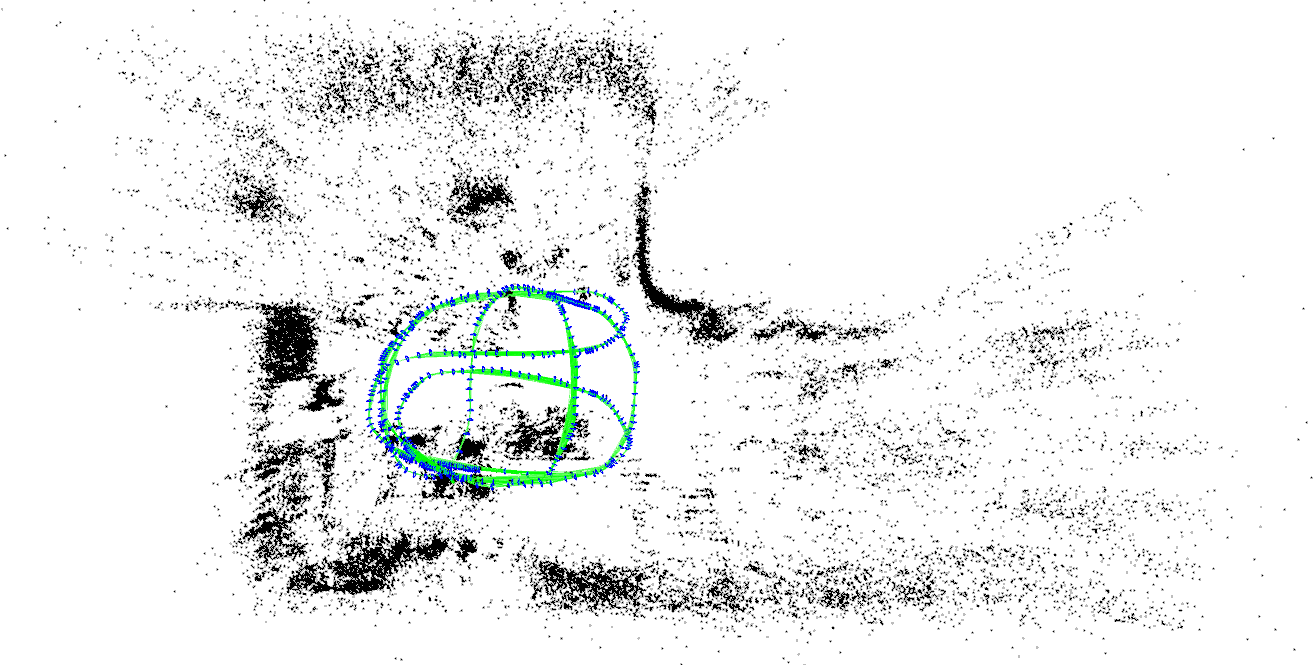}
		\captionsetup{justification=centering}
		\label{fig:intro_b}
	\end{subfigure} \\ \vspace{0pt}
	\begin{subfigure}[t]{0.48\textwidth}
		\includegraphics[width=0.995\linewidth, height=0.38\linewidth]{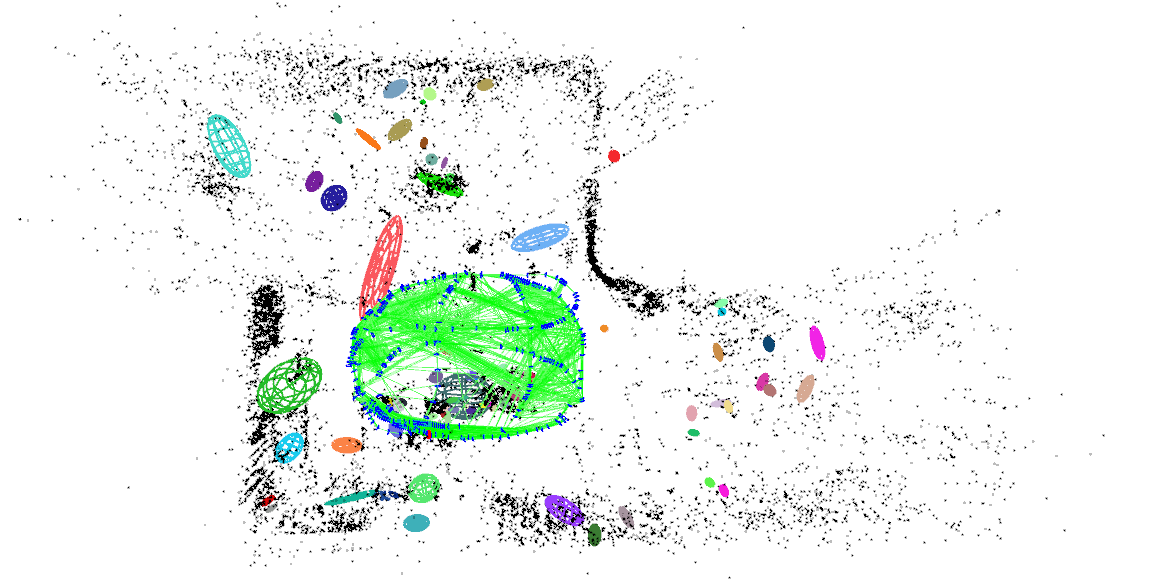}
		\captionsetup{justification=centering}
		\label{fig:intro_c}
	\end{subfigure}\vspace{-7pt}
	\caption{Top: Groundtruth map. Middle: ORB SLAM2 with loop closure disabled. Bottom: VOOM. The colored ellipsoids represent the 3D objects, blue cones illustrate the camera poses, and green lines represent covisibility relationships between keyframes. The map generated by ORB-SLAM2 has redundant point clouds and blurry boundaries, while our VOOM builds a more lightweight yet semantic enhanced map, resulting in more accurate camera pose estimation.}
	\label{fig:intro}
      \vspace{-1mm}
\end{figure}


However, traditional visual odometry or SLAM frameworks that use map points as landmarks lack high-level information, resulting in limited robustness and applicability~\cite{engel2014lsd, mur2017orb, qin2018vins}, which has led to a growing interest in object-level odometry and mapping algorithms~\cite{Runz2018MaskFusion, Mccormac2018Fusion, rosinol2020kimera}, especially in some lightweight approaches using cuboid or dual quadrics as landmarks~\cite{Yang2018CubeSLAM, Rubino2018}. Current algorithms mainly focus on constructing more accurate object landmarks~\cite{liao2022so, lin2021robust}, and improving localization accuracy using object information~\cite{nicholson2018quadricslam, liao2020rgb, wang2023qiso}. Although these methods incorporate object residuals into bundle adjustment, they rely on an accurate initial guess of camera poses from feature-based techniques such as ORB-SLAM2~\cite{mur2017orb}. 
Their localization results are similar to or sometimes worse than feature-based methods. This indicates that current object-based methods have not fully utilized valuable object information to improve localization accuracy. How to further exploit object information to improve visual SLAM results remains challenging.

In this paper, we propose a novel Visual Object Odometry and Mapping framework, named VOOM, which applies both low-level feature points and high-level objects to the entire process of odometry and mapping, as shown in Fig. \ref{fig:intro}. In odometry, we combine a residual model of dual quadrics based on the Wasserstein distance~~\cite{wang2021normalized} with an object data association method based on the normalized Wasserstein distance to construct an accurate object-level map. By utilizing the object-level information, the current frame's keypoints are better matched with the previous map points, which results in an enhanced pose estimation for the frame. In terms of visual object mapping, we establish an object-based covisibility graph based on the observation relationships between keyframes and objects. The map points corresponding to the current neighbors in the graph are updated, thereby updating the point-based covisibility graph and the local map. Through a bundle adjustment, the poses of local keyframes and the positions of map points are optimized. In experiments, we have demonstrated that the proposed VOOM with only odometry and mapping components even outperforms the ORB-SLAM2 with loop closure in most sequences. To the best of our knowledge, this is the first paper to implement and show that object-oriented SLAM systems using dual quadrics and feature points can achieve better localization accuracy than state-of-the-art feature-based visual SLAM systems.

To sum up, our main contributions are threefold: 
\begin{itemize}
	\item the development of a novel visual object odometry and mapping framework using both feature points and dual quadrics as landmarks;
	\item the design of effective algorithms for object optimization, object association, and objects-based map points association for constructing the map with hierarchical landmarks; 
	\item extensive experimental validation demonstrating the superiority of the proposed approach compared to state-of-the-art methods.
\end{itemize}

\section{Related Work}
\label{literatures}
\subsection{Visual Odometry and Mapping}
Visual Odometry and mapping are often classified into direct and feature-based methods, depending on how they extract information from images to estimate camera motion and reconstruct maps. 

Direct methods, such as LSD-SLAM~\cite{engel2014lsd} and Direct Sparse Odometry (DSO)~\cite{engel2017direct}, estimate camera motion and reconstruct maps by directly utilizing pixel intensity information from images.  Direct Sparse Mapping (DSM)~\cite{zubizarreta2020direct} extends DSO by reusing existing map information. Although the direct methods can handle low-textured and blurred images, they are sensitive to camera parameters and exposure. 

In contrast, feature-based methods are more robust in variable lighting conditions by tracking extracted features, such as SIFT, BRIEF, and ORB. In early visual SLAM, filtering-based methods were widely used~\cite {davison2007monoslam, civera20101} for tracking features thanks to its low computational cost. However, with the improvement of computing power, systems like PTAM~\cite{klein2007parallel} began to use bundle adjustment to optimize keyframe poses and divide the whole process into tracking and mapping, which is more accurate than filtering methods. Based on these ideas, the monocular ORB-SLAM~\cite{mur2015orb} tracks the more robust ORB features and limits the complexity of pose optimization and mapping by building a covisibility graph. ORB-SLAM2~\cite{mur2017orb} further extends the system for stereo and RGB-D cameras.
ORB-SLAM2 has been proven to be versatile, efficient, and relatively accurate in practice. Therefore, in our work, we adopt their methods for points-level processing. Building upon this foundation, we introduce the object-level landmarks and seamlessly integrate them into the entire odometry and mapping process.

\subsection{Object-oriented SLAM}
Semantic SLAM can represent a high-level understanding of complex environments. Among different visual semantic SLAM approaches, object-oriented SLAM offers a lightweight solution that can enhance localization and mapping performance by leveraging object relations and geometric constraints.

SLAM\texttt{++}~\cite{Moreno2013SLAM} is an early proposed object-oriented mapping system, which shows its superiority in map capacity compared to dense semantic mapping without sacrificing descriptiveness. SLAM\texttt{++} is not versatile, as it can only track domain-specific objects with known geometric shapes. In ~\cite{Sunderhauf2017mean}, arbitrary objects are built as point cloud models using bounding box detection and 3D segmentation. Based on these mapping systems, Fusion\texttt{++}~\cite{Mccormac2018Fusion} further explores the utilization of reconstructed objects to enhance the precision of localization. Using only the 3D object information stored in a 6DoF pose graph, Fusion\texttt{++} has achieved better accuracy.
With the advent of CubeSLAM~\cite{Yang2018CubeSLAM} and QuadricSLAM~\cite{nicholson2018quadricslam}, researchers are motivated to use compact geometry shapes to represent the objects, instead of providing a detailed description of the object's surface. Compared with using cuboids used in ~\cite{Yang2018CubeSLAM}, using dual quadrics in object-oriented SLAM has the superiority of mathematical completeness in projective geometry~\cite{Rubino2018}. Some variations of quadrics-based approaches focus on constructing precise observation models for mapping~\cite{Wu2020EAO, wu2023object, lin2021robust, liao2022so}. Other researchers have also attempted to improve the localization performance by adding objects residual into the cost function of optimization~\cite{nicholson2018quadricslam, liao2020rgb, wang2023qiso}, but there is only negligible improvement over feature-based methods. OA-SLAM~\cite{zins2022oa} has proposed an object-aided relocalization approach, which uses an object model for coarse pose estimate and points-based methods for refining the localization performance. It is indicated in OA-SLAM that objects excel at providing high-level semantic information and having strong discriminating power, whereas points offer greater accuracy in spatial localization. 

Inspired by the coarse-to-fine idea of OA-SLAM, we propose a more logical strategy for enhancing odometry and mapping. Instead of directly introducing object residuals into optimization, we tightly incorporate 3D objects as landmarks to assist feature point associations. This enables us to utilize 3D map points repeatedly, not just for temporally adjacent frames but also for spatially adjacent ones. As a result, we can reduce the accumulation of errors in odometry.

\begin{figure*}[htbp]
	\centering
	\includegraphics[
width=0.9\linewidth]{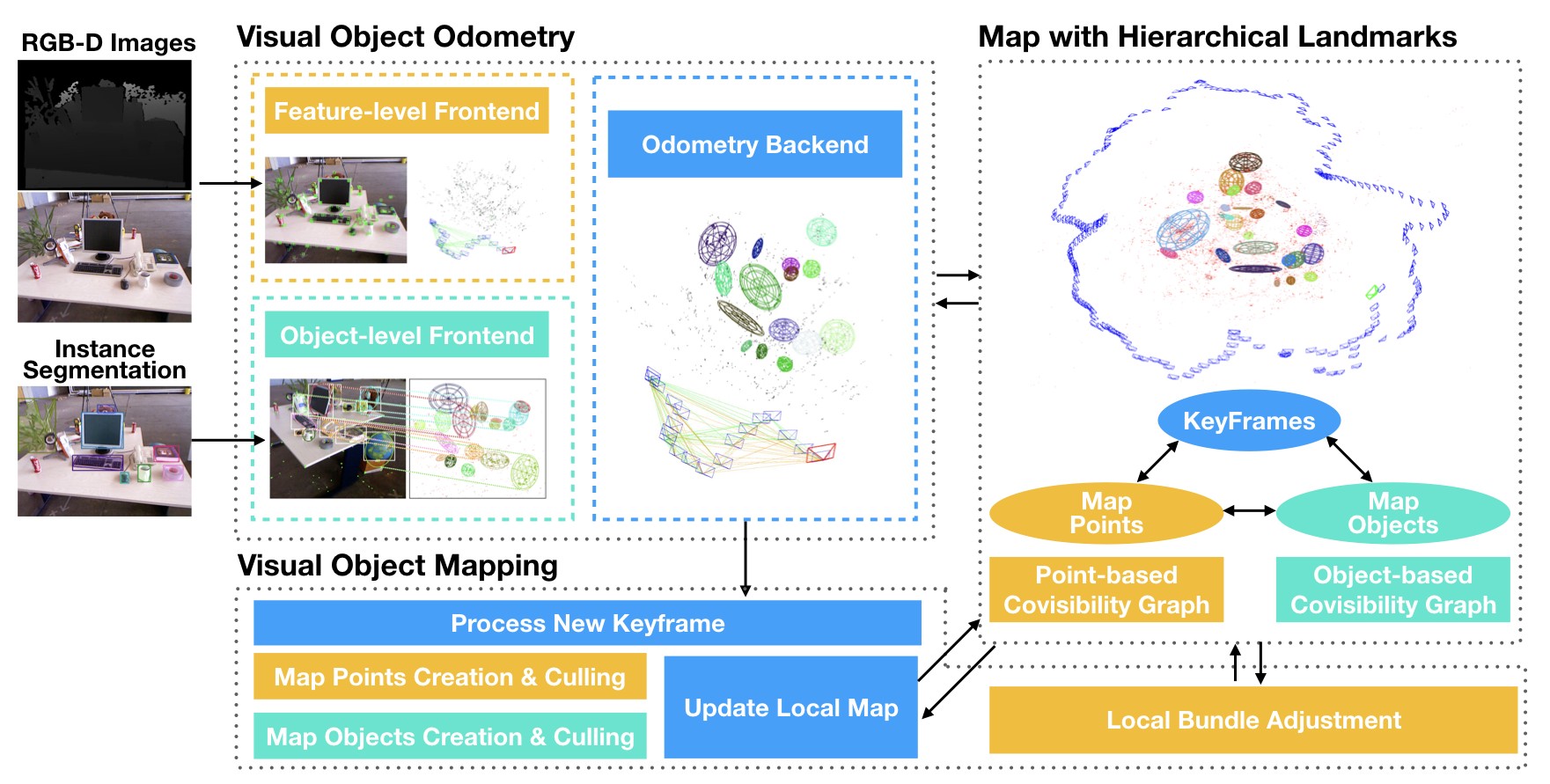}
	\caption{The proposed VOOM framework uses both low-level feature-based map points and high-level objects as landmarks. The color yellow indicates the process has only feature points participated, the color mint reveals that it is a pure object-based process, and the color blue means both landmarks contribute to the procedure.}
	\label{fig:pipeline}
\end{figure*}

\section{VOOM}
\label{method}

\subsection{System Overview}

The proposed VOOM takes an RGB-D image as input and outputs a 3D map composed of map points and 3D objects. The overall process is divided into two main parts: Visual Object Odometry and Visual Object Mapping, as shown in Fig.\ref{fig:pipeline}. In the odometry part, the frontend processes feature points and information from instance segmentation results implemented by YOLOv8~\cite{yolov8}. We use the method in ORB-SLAM2 to process ORB feature points and pose prediction. In the object-level frontend, we estimate ellipses using the direct least squares method~\cite{fitzgibbon1999direct} based on a set of counter points obtained from instance segmentation. We determine whether objects are associated based on the Wasserstein distance between the estimated ellipse and the projection of the 3D ellipsoid onto the image plane, which we present in Sec.~\ref{obs_model} and \ref{da_subsec} in detail. In the odometry backend, we use associated objects to update the data association of map points based on the optimized frame pose, introduced in the first part of Sec.~\ref{oam}. Once the frame is determined as a keyframe, it is passed to the Visual Object Mapping thread, introduced in the second part of Sec.~\ref{oam}.



\subsection{Observation Model of Objects}
\label{obs_model}
We perform instance segmentation to detect objects in each incoming image. 
It has been demonstrated in QISO-SLAM~\cite{wang2023qiso} that using contours of instance segmentation can achieve higher accuracy on object-oriented mapping and localization than using bounding boxes. In line with that, we estimate an observation ellipse $E^\text{obs}_{fk}$ of a $k$-th object on $f$-th frame by minimizing the sum distance between each segmentation contour point and the ellipse outline.

The object can be represented by a constrained dual quadric $\mathbf{Q}_k^\ast\in \mathbb{R}^{4\times4}$. It is projected onto the image plane of $f$-th frame as a dual conic:
\begin{equation}
\label{eqa_cam_proj}
\mathbf{C}_{fk}^\ast = \mathbf{P}_f\mathbf{Q}_k^\ast\mathbf{P}_f^\top\ ,
\end{equation}
where $\mathbf{P}_f = \mathbf{K}_f[\mathbf{R}_f|\mathbf{t}_f^\top]\in \mathbb{R}^{3\times4}$ is the camera projection matrix that contains the matrix of the intrinsic parameters $\mathbf{K}_f\in \mathbb{R}^{3\times3}$, rotation matrix $\mathbf{R}_f\in \mathbb{R}^{3\times3}$ and translation vector $\mathbf{t}_f^\top\in \mathbb{R}^{3\times1}$. $\mathbf{C}_{fk}^\ast$ represents an estimated ellipse $E^{\text{est}}_{fk}$. An error function can thus be constructed with extracted ellipses of $E^{\text{obs}}_{fk}$ and $E^{\text{est}}_{fk}$.

Inspired by OA-SLAM~\cite{zins2022oa}, we also utilize a 2D Gaussian distributions $\mathcal{N}(\boldsymbol{\mu}, \boldsymbol{\Sigma}^{-1})$ to interpret the ellipses: 
\begin{equation}
\boldsymbol{\mu} = \left( \begin{array}{cc}
p_x  \\ p_y
\end{array}\right)
\ , \ 
\mathbf{\Sigma}^{-1} = \boldsymbol{R}^\top(\theta)\left(
\begin{array}{cc}
\frac{1}{\alpha^2} & 0   \\
0 & \frac{1}{\beta^2} 
\end{array} 
\right)\boldsymbol{R}(\theta)\ ,
\end{equation}
where $[p_x, p_y]\top$, $[\alpha, \beta]\top$ and $\theta$ describes the center point, the axes length, and the rotation angle of an ellipse, respectively. The error between $E^{\text{obs}}_{fk}$ and $E^{\text{est}}_{fk}$ is measured by the $2$nd order Wasserstein distance between two Gaussian distributions $\mathcal{N}_{fk}^{\text{obs}}(\boldsymbol{\mu_1}, \boldsymbol{\Sigma_1}^{-1})$ and $\mathcal{N}_{fk}^{\text{est}}(\boldsymbol{\mu_2}, \boldsymbol{\Sigma_2}^{-1})$ is computed by:

\begin{equation}
\text{W}_2^2(\mathcal{N}_{fk}^{\text{obs}}, \mathcal{N}_{fk}^{\text{est}}) = \lVert\boldsymbol{\mu_1} - \boldsymbol{\mu_2}\rVert^2_2 \  + \lVert \boldsymbol{\Sigma}_1^{\frac{1}{2}}- \boldsymbol{\Sigma}_2^{\frac{1}{2}} \rVert ^2_\text{F} \ ,
\label{wasser}
\end{equation}
where $\lVert\cdot\rVert_\text{F}$ is the Frobenius norm. When the $k$-th object is observed by a set $\mathcal{F}_L$ of frames, the parameters of the corresponding ellipsoid are estimated by minimizing: 
\begin{equation}
J = \sum_{f\in\mathcal{F}_L} \text{W}_2^2(\mathcal{N}_{fk}^{\text{obs}}, \mathcal{N}_{fk}^{\text{est}})\ .
\end{equation}

\begin{figure}[htbp]
	\begin{subfigure}[t]{0.24\textwidth}
		\includegraphics[scale=0.14]{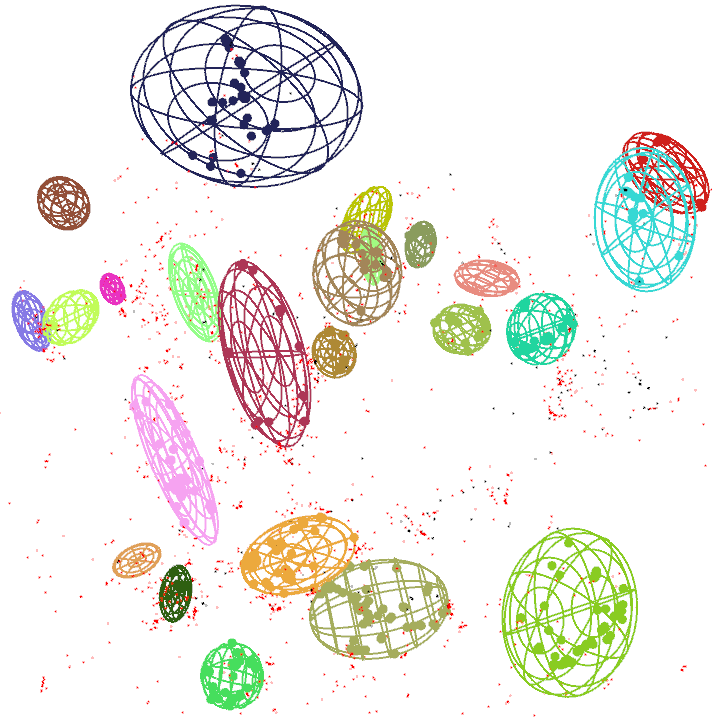}
		\caption{3D Map}
		\label{map_3d}
	\end{subfigure} \noindent  \hspace{-4.5mm}
	\begin{subfigure}[t]{0.24\textwidth}
		\includegraphics[scale=0.20]{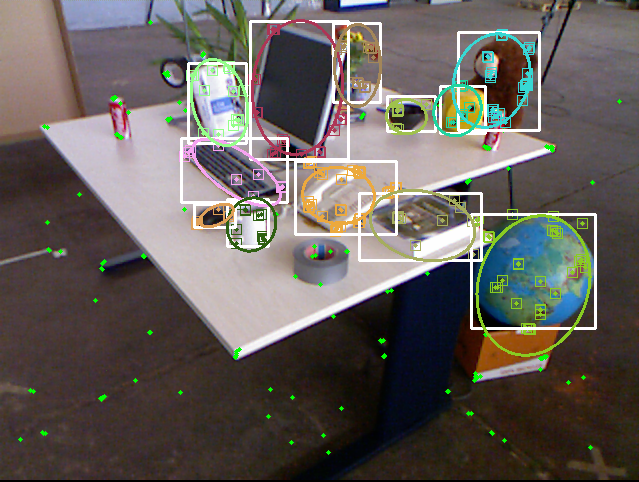}
		\caption{Image plane}
		\label{map_im}
	\end{subfigure}
	\caption{Illustration of associating map points based on the associated objects.}
	\label{fig:mapp}
\end{figure}

\subsection{Object-level Data Association}
\label{da_subsec}
In previous works on objected-oriented SLAM based on dual quadrics, the Intersection over Union (IoU) between reprojected boxes and detected bounding boxes was used as the criterion for object data association. However, there are multiple issues by using such a criterion. First, a bounding box enclosing an ellipse has a larger area than the ellipse itself, which increases the overlap among irrelevant objects. Second, we observe that the data association performance of tiny objects is severely affected by camera pose error, as the IoU-based metrics are highly sensitive to it. As we have employed Wasserstein distance to construct the residual model, it can also serve as a measurement for object data association. Normalized Wasserstein distance was proposed in ~\cite{wang2021normalized} for tiny object detection. Based on \eqref{wasser}, the normalized Wasserstein distance is written as:
\begin{equation}
\label{norm_wasser}
W_{\text{n}}(\mathcal{N}_{fk}^{\text{est}}, \mathcal{N}_{fk}^{\text{obs}}) = \exp\left(-\frac{\sqrt{W^2_2(\mathcal{N}_{fk}^{\text{est}}, \mathcal{N}_{fk}^{\text{obs}})}}{C}\right) \ .
\end{equation}
Here, $C$ is a constant. This metric incorporates the center point, length axes, and orientation information of the ellipses, which is more comprehensive than IoU. 
\subsection{Coarse-to-fine Odometry and Mapping}
\label{oam}
We present our method for visual object odometry and mapping in a coarse-to-fine manner. The main idea is to use the robustness of 3D object-level landmarks to enhance the efficiency and accuracy of feature-based map points to improve the odometry and mapping performance. 

\subsubsection{Coarse-to-fine Visual Object Odometry Backend}
Given a coarse pose predicted by the motion model from the feature-level frontend and the associated objects from the object-level frontend, we search the matches between all map points corresponding to each object (marked as the colored large points inside each ellipsoid in Fig. \ref{map_3d}) and extracted feature keypoints belonging to the associated instance (marked as the colored rectangles inside each bounding box in Fig. \ref{map_im}) by distance of ORB descriptors~\cite{mur2015orb}. This allows us to achieve more efficient and precise map points association by narrowing down the search area, and to retrieve more map points from earlier frames, rather than from recent ones. With the matches between 3D map points and keypoints, the frame pose can be refined for the first time by minimizing the reprojection error. Next, we update the local map for the current frame by searching the previous keyframes that observe the associated map points and all map points that are related to those keyframes to match the current keypoints. Then we perform a finer pose optimization for the second time to obtain the current frame pose. Using the same criteria as in ORB-SLAM2, a keyframe is decided to be created and passed to the visual object mapping thread. 

\subsubsection{Coarse-to-fine Visual Object Mapping}
\begin{figure}[t]
	\centering
	\begin{subfigure}[t]{0.24\textwidth}
		\includegraphics[width=0.98\linewidth]{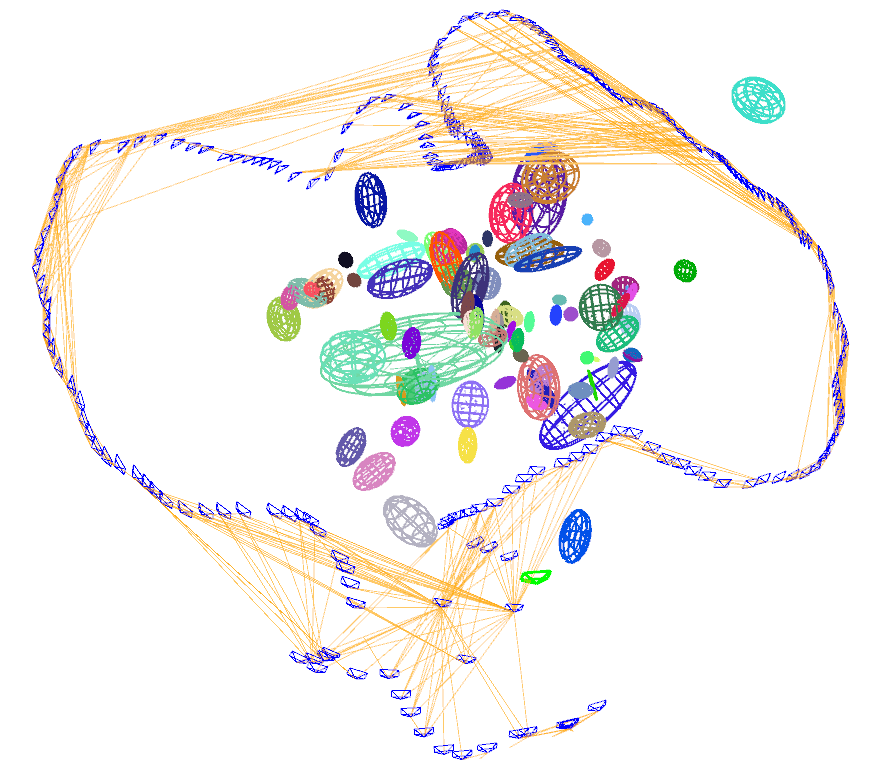}
		\caption{Objects-based graph}
		\label{fig:covisi_b}
	\end{subfigure} \noindent  \hspace{-1.5mm}
	\begin{subfigure}[t]{0.24\textwidth}
		\includegraphics[width=0.98\linewidth]{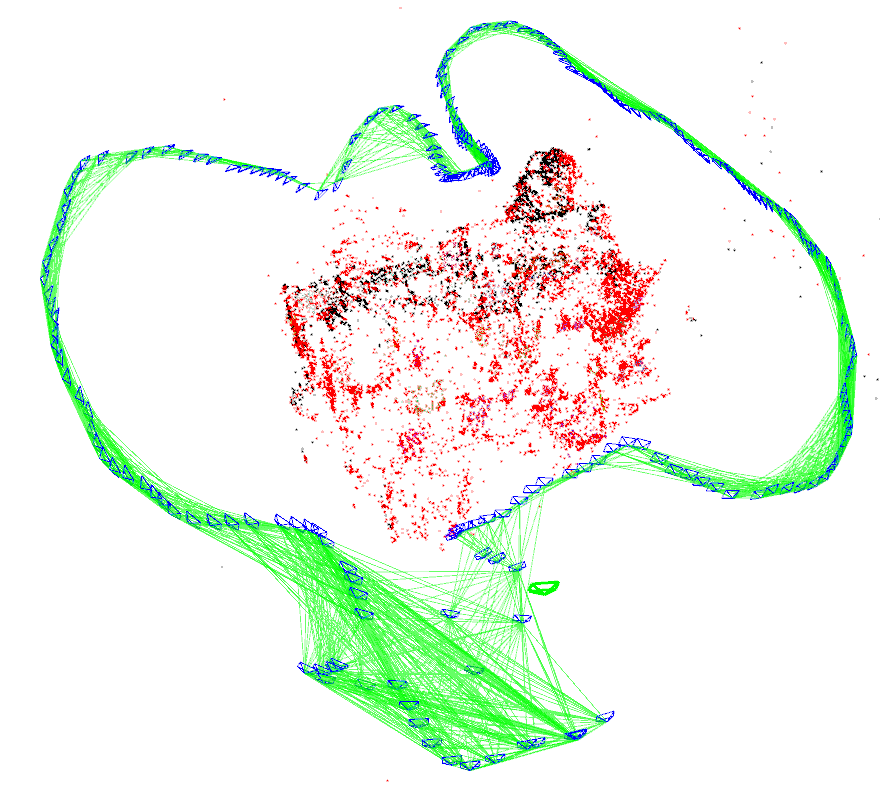}
		\caption{Points-based graph}
		\label{fig:covisi_a}
	\end{subfigure} 
	\caption{Illustration of covisiblity graphs.}
\end{figure}

After obtaining a reliable pose estimation and map points association for the current keyframe, we update the local map based on the object covisibility graph in visual object mapping. The object covisibility graph records the covisibility relations between different objects in different keyframes, as is shown in Fig. \ref{fig:covisi_b}. It can help us select a subset of keyframes most relevant to the current frame. In contrast to the traditional covisibility graph based on map points (illustrated in Fig. \ref{fig:covisi_a}), our object covisibility graph has a smaller data size and higher robustness. With the object-based covisibility graph, we can retrieve earlier keyframes that have previously observed objects to construct a local map for local bundle adjustment. Due to the robustness of objects, it implicitly achieves some of the effects of loop detection, even with only local mapping. In local BA, we use the retrieved keyframes and the corresponding map points, same as in ~\cite{mur2017orb}. Residuals of objects do not participate in BA because the object model itself is less accurate than the feature-based one, as shown in many previous works~\cite{nicholson2018quadricslam, liao2020rgb, wang2023qiso}. After BA, the keyframe poses and positions of map points are refined, and outliers are processed. The parameters of each object are optimized according to the poses of its observed keyframes separately.

\section{Experimental Results}
\label{experiments}
In this section, we present the experimental results of our proposed method. Our systems take the results of instance segmentation using YOLOv8~\cite{yolov8} as input, which is processed offline by Python. We only keep the detection with a confidence higher than $0.2$ in the experiments. The online system is built upon the open-source C++ system ORB-SLAM2 with loop closure disabled. We demonstrate the localization performance in subsection \ref{subsec_res_loc}, the object data association results in subsection \ref{subsec_res_obj}, and an ablation study in \ref{subsec_res_abl} on TUM RGB-D dataset~\cite{sturm12iros} and LM-Data~\cite{Characterizing19}.

\subsection{Localization}
\label{subsec_res_loc}
\begingroup
\renewcommand{\arraystretch}{1.1}
\setlength{\tabcolsep}{7.5pt}
\begin{table*}[thbp]
	\centering
	\caption{Localization RMSE comparison on TUM sequences.} \label{table_tum_tracking} 
	\begin{tabular}{lcccccccc} \toprule
		\textbf{Sequence} & \textbf{Length} & \textbf{ORB-SLAM2}  &  \makecell[c]{\textbf{ORB-SLAM2}\\\textbf{No LC}}  & \textbf{Fusion++} & \textbf{QuadricSLAM}  & \textbf{RGB-D}~\cite{liao2020rgb} & \textbf{QISO-SLAM }     & \textbf{VOOM}  \\
		\midrule
		Fr1\_desk1 & 9.263\SI{}{m}  & \textbf{0.0151} & 0.0193 & 0.049 & 0.0167  & \underline{0.0156}   & 0.0166   & 0.0189    \\
		Fr1\_desk2  & 10.161\SI{}{m}  & 0.0243 & 0.0264 & 0.153 & 0.0245     & \textbf{0.0210} &  \underline{0.0220}   & 0.0245     \\
		Fr2\_desk   & 18.880\SI{}{m} & \underline{0.0087}   & 0.0114 & 0.114 & 0.0124 & 0.0100   &  0.0099    & \textbf{0.0080}      \\
		Fr2\_person  & 17.044\SI{}{m} & \underline{0.0065}   & 0.0866   & - & - & -   & -     & \textbf{0.0062}     \\
		Fr3\_teddy & 19.807\SI{}{m} & \underline{0.0370}  & 0.0417  & - & - & -   & -     & \textbf{0.0179}  \\
        Fr3\_office & 21.455\SI{}{m} & \underline{0.0107}  & 0.0113  & 0.108 & 0.0230 & 0.0111  & 0.0112    & \textbf{0.0100}  \\
		\bottomrule
	\end{tabular} \\
     \begin{tablenotes} 
     \item The best result is highlighted in \textbf{bold}, and the second best is \underline{underlined}. 
     \end{tablenotes}
 \label{loc_tum}
\end{table*}
\endgroup

We conducted experiments on two datasets to compare the localization accuracy of our proposed method with several state-of-the-art methods. To quantify the localization accuracy, we use the root-mean-square error (RMSE) as the evaluation metric. We compare VOOM with ORB-SLAM2~\cite{mur2017orb}, ORB-SLAM2 No LC, Fusion++~\cite{Mccormac2018Fusion} QuadricSLAM~\cite{nicholson2018quadricslam}, a RGB-D object SLAM~\cite{liao2020rgb} and QISO-SLAM~\cite{wang2023qiso} on TUM RGB-D Dataset, while we only compare with ORB-SLAM2 and ORB-SLAM2 No LC on LM-Data. ORB-SLAM2 is a representative feature-based SLAM method, and ORB-SLAM2 No LC is the same method with loop closure detection disabled. The other four frameworks are object-oriented SLAM, which attempted to improve the localization accuracy results using objects. Among them, Fusion++ uses refined dense object models to improve the TSDF odometry without traditional feature-based tracking. QuadricSLAM, RGB-D~\cite{liao2020rgb}, and QISO-SLAM rely on the ORB-SLAM2 results as the initial camera pose and only use object information for refinement. In contrast, our method fuses feature points and object information in a unified framework to obtain the localization results directly. 

Table \ref{loc_tum} presents the localization performance on five data sequences: Fr1\_desk1, Fr1\_desk2, Fr2\_desk, Fr2\_person, and Fr3\_office. The first column lists the sequence names, the second column shows the total lengths of the sequences, and the subsequent columns report the RMSE values obtained by each algorithm. The table reveals that our algorithm outperforms the others, especially in three longer sequences, and consistently improves upon ORB-SLAM2 No LC across all sequences. The other three object-oriented methods rely on the results of ORB-SLAM2 with loop closure. Still, they do not exhibit noticeable enhancement in localization accuracy, which is also analyzed and demonstrated in their papers. Another remarkable observation is that on the Fr2\_person sequence with several dynamic objects, our localization result achieves a $92.8\%$ error reduction over ORB-SLAM2 No LC. This further demonstrates the potential of our work to make contributions to dynamic SLAM.

\begin{figure}[t]
	\centering
	\begin{subfigure}[t]{0.48\textwidth}
		\includegraphics[width=0.98\linewidth, height=0.41\linewidth]{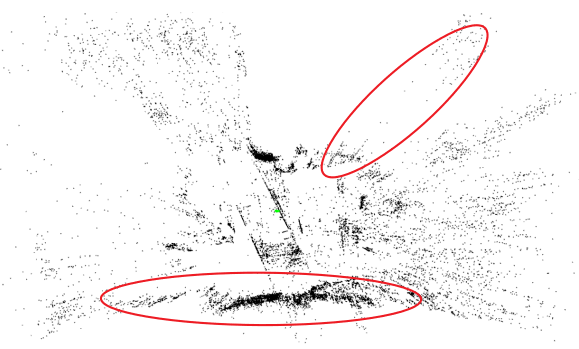}
        \vspace{0pt}
		\captionsetup{justification=centering}
		\caption{ORB-SLAM2}
		\label{fig:loc_map_b}
	\end{subfigure} \\  \vspace{1mm}
	\begin{subfigure}[t]{0.48\textwidth}
		\includegraphics[width=0.98\linewidth, height=0.41\linewidth]{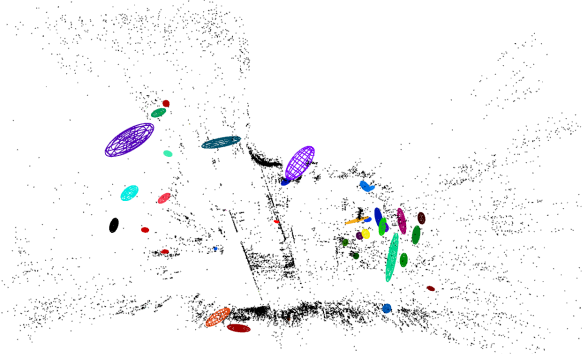}
        \vspace{-1pt}
		\captionsetup{justification=centering}
		\caption{VOOM}
		\label{fig:loc_map_d}
	\end{subfigure}
    \vspace{0mm}
	\caption{The qualitative mapping results on the Diamond Ground Sequence. The obvious distorted parts of the ORB-SLAM2 map are circled with red color.}
	\label{fig:loc_map_ground}
      \vspace{0mm}
\end{figure}

Table \ref{loc_diamond} reports the results on four LM-Data Diamond sequences: Ground, Walk Head, Walk, and MAV. VOOM exceeds ORB-SLAM2 and ORB-SLAM2 No LC in all sequences. A notable result is obtained on the Diamond Walk sequence, where the camera travels long and revisits some same places. The error of ORB-SLAM2 without loop closure detection is over \SI{5}{\cm}, and VOOM declined it by about $3/4$. The results above prove that our proposed approach can effectively reduce the odometry drift without loop closure detection, as it utilizes both the robustness of objects and the accuracy of features. Furthermore, because VOOM updates and optimizes local maps with objects and feature points throughout the whole process, unlike loop closure which only optimizes when loops are detected, it even surpasses ORB-SLAM2 with loop closure. With the qualitative mapping results on the Diamond Ground sequence displayed in Fig. \ref{fig:loc_map_ground}, we can better understand how VOOM improves the performance\footnote{Videos of this and other experiments can be found at \url{https://www.bilibili.com/video/BV1w14y1C7Jb/}}. It is observed that ORB-SLAM2 cannot close the loop due to the limited view range of the ground robot, resulting in a redundant and distorted point cloud map. Nonetheless, our system is capable of correcting the distortion and building a more structured map when the camera re-observes certain objects.

\begingroup
\renewcommand{\arraystretch}{1.1}
\begin{table}[t]
	\centering
	\caption{Localization RMSE comparison on LM-Data Diamond sequences. The best result is highlighted in \textbf{bold}.} \label{table_tum_tracking} 
	\begin{tabular}{ccccc} \toprule
		\textbf{Sequence} & \textbf{Length} & \textbf{ORB-SLAM2}  &  \makecell[c]{\textbf{ORB-SLAM2}\\\textbf{No LC}}   & \textbf{VOOM}  \\
		\midrule
		Ground & 5.87\SI{}{m}  & 0.1658   & 0.1665   & \textbf{0.1287}   \\
		Walk Head   & 13.82\SI{}{m} & 0.0135  &  0.0193    & \textbf{0.0125}    \\
		Walk  & 22.56\SI{}{m}    & 0.0213 &  0.0596   & \textbf{0.0150}   \\
		MAV  & 33.84\SI{}{m}  & 0.0132  & 0.0195   & \textbf{0.0114}    \\
		\bottomrule
	\end{tabular}
	\label{loc_diamond}
\end{table}
\endgroup

\subsection{Object Data Association Results}
\label{subsec_res_obj}
Both the observation model and object association influence the object association results. Thus, we evaluated and compared four object association methods and two observation models. The metric we used was the total number of reconstructed objects, reflecting the object-level map accuracy. The closer the number is to the ground truth, the better the data association performance is. Unlike many previous works, such as QuadricSLAM~\cite{nicholson2018quadricslam}, OA-SLAM~\cite{zins2022oa}, and QISO-SLAM~\cite{wang2023qiso}, we kept all the 3D objects added to the map without culling objects with low occurrence frequency to demonstrate the real effect of different methods in object data association. We used three sequences from two datasets: Fr2\_desk from TUM RGB-D, Diamond Walk (D\_Walk), and Diamond MAV (D\_MAV) from LM-Data. 

Table \ref{da1} summarizes the results. We compare four object data association methods: DA1, DA2, DA3, DA4. Among them, DA1 is the method used in OA-SLAM~\cite{zins2022oa}, DA2 uses IoU as metric, DA3 is the method used in QISO-SLAM2~\cite{wang2023qiso}, and DA4 is our proposed approach. The main metric for DA1-DA3 is IoU, with a basic threshold of $0.3$. The difference is that DA1 only allows objects of the same category to be associated, while DA2 and D3 associate objects of any category. Moreover, DA3 also considers shape similarity and lowers the IoU threshold to $0.1$ when the object labels are the same. As for our proposed DA4, We empirically set the constant $C$ in \eqref{norm_wasser} as $C=10$ and a very low threshold of $0.005$ as we found it is not as sensitive as the IoU threshold. We also compare two different observation models: M1 and M2. M1 is the model used in OA-SLAM, and M2 is our proposed one. As can be seen, our proposed DA4 outperforms the other methods in most cases. It is observed that DA1 has the most number of redundant objects among all data association methods. This is because it is very common for an object to be classified into different categories in real cases. DA3 performs better than DA2, as it incorporates more criteria for object data association. As for the observation model, M2 shows better performance than M1 overall. Particularly on Fr2\_desk and Diamond Walk sequences, all data related to M2 are superior to M1. The only exception is the Diamond MAV sequence, where M2 is slightly inferior to M1. It may be because objects are estimated with larger sizes using M1, which helps object association in faster-speed scenarios. In general, the combination of DA4 and M2, which we proposed, is suitable for most situations, while in some cases, using DA4 and M1 can also achieve sufficient results.

\begingroup
\renewcommand{\arraystretch}{1.1}
\setlength{\tabcolsep}{5.2pt}
\begin{table}[t]
	\centering
	\caption{Object Data Association Results.}
	\begin{tabular}{ccc|cc|cc|cc|c} \toprule
		\multirow{2}{*}{\textbf{Dataset}}  & \multicolumn{2}{c|}{\textbf{DA1}}    & \multicolumn{2}{c|}{\textbf{DA2}}  & \multicolumn{2}{c|}{\textbf{DA3}}   & \multicolumn{2}{c|}{\textbf{DA4 (Ours)}}  & \multirow{2}{*}{\textbf{GT}}       \\ \cline{2-9} 
		& \begin{tabular}[c]{@{}c@{}}\textbf{M1}\end{tabular}     & \begin{tabular}[c]{@{}c@{}}\textbf{M2}\end{tabular} & \begin{tabular}[c]{@{}c@{}}\textbf{M1}\end{tabular}     & \begin{tabular}[c]{@{}c@{}}\textbf{M2}\end{tabular}   & \begin{tabular}[c]{@{}c@{}}\textbf{M1}\end{tabular}     & \begin{tabular}[c]{@{}c@{}}\textbf{M2}\end{tabular}    & \begin{tabular}[c]{@{}c@{}}\textbf{M1}\end{tabular}     & \begin{tabular}[c]{@{}c@{}}\textbf{M2}\end{tabular} &   \\
		\midrule 
		Fr2\_desk & 131 & 119  & 68   & 64  & 55 & 54  & 52  & \textbf{49} & 44  \\ 
		\makecell[c]{D\_Walk} & 73 & 74  & 71   & 59  & 58 & 55  & 59  & \textbf{53} & 50  \\ 
		\makecell[c]{D\_MAV} & 74 & 80  & 73   & 78  & 51 & 59  & \textbf{50}  & 52 & 43  \\ 
		\bottomrule
	\end{tabular}
	\label{da1}
\end{table}
\endgroup

\subsection{Ablation Study}
\label{subsec_res_abl}
We conducted an ablation study to evaluate the contribution of different parts of VOOM. Besides the full VOOM (F), we also show the results of three variants: VOOM (O), VOOM (M), and VOOM (A). VOOM (O) represents that we use objects only in odometry, while VOOM (M) means objects only contribute to mapping. In VOOM (A) we replaced the proposed data association method and observation model with the ones used in OA-SLAM to demonstrate their impact on localization performance.  

The RMSE of the ablation study is presented in Table \ref{loc_ablation}. We chose Fr2\_desk and Fr2\_person from the TUM RGB-D dataset, and Diamond Walk from the LM-Data. The baseline is ORB-SLAM2 No LC, which is the basis of our algorithm. First, all variants surpass the original ORB-SLAM2 No LC, which demonstrates the effectiveness of each improvement in our framework. Furthermore, the results show that the full VOOM algorithm achieves the best performance, followed by VOOM (M), which indicates that local bundle adjustment with objects contributes slightly more than pose estimation in odometry. Additionally, VOOM (A) is worse than the full VOOM, which suggests that the superior data association method and observation model we proposed can also enhance localization accuracy.

\begingroup
\setlength{\tabcolsep}{5pt}
\renewcommand{\arraystretch}{1.1}
\begin{table}[t]
	\centering
	\caption{Ablation Study. The best result is highlighted in \textbf{bold}, and the second best is \underline{underlined}.} \label{table_tum_tracking} 
	\begin{tabular}{cccccc} \toprule
		\textbf{Dataset} & \makecell[c]{\textbf{ORB}\\\textbf{No LC}} & \makecell[c]{\textbf{VOOM} \\ \textbf{(O)}}  &  \makecell[c]{\textbf{VOOM} \\ \textbf{(M)}} & \makecell[c]{\textbf{VOOM} \\ \textbf{(A)}} & \makecell[c]{\textbf{VOOM} \\ \textbf{(F)}}  \\
		\midrule \vspace{4pt}
		TUM Fr2\_desk & 0.0114  & 0.0086   & \underline{0.0084} & 0.0105   & \textbf{0.0080}   \\ \vspace{4pt}
		TUM Fr2\_person & 0.0866  & 0.0065   & \underline{0.0063} & 0.0071   & \textbf{0.0062}   \\ 
		Diamond  Walk   & 0.0596 & 0.0205  &  \underline{0.0172}  & 0.0249   & \textbf{0.0150}    \\
		\bottomrule
	\end{tabular}
	\label{loc_ablation}
\end{table}
\endgroup

\section{Conclusion}
\label{conclusion}
Our experimental results have demonstrated that by incorporating object information, VOOM is able to achieve better localization performance than the state-of-the-art feature-based visual SLAM methods. This level of improvement has not been previously observed in object-oriented SLAM approaches using dual quadrics as landmarks, to the best of our knowledge. Our system has the odometry and mapping components, with a primary focus on short-term and mid-term performance. In future work, we plan to build upon this foundation by introducing a loop closure and relocalization algorithm that integrates both objects and points efficiently, with the goal of achieving improved long-term SLAM performance.

\addtolength{\textheight}{0cm}   

\bibliographystyle{IEEEtran}
\bibliography{wang2024voom}

\end{document}